\documentclass{article}
\usepackage{ijcai21}

\usepackage{times}
\usepackage{soul}
\usepackage{url}
\usepackage[hidelinks]{hyperref}
\usepackage[utf8]{inputenc}
\usepackage[small]{caption}
\usepackage{graphicx}
\usepackage{amsmath}
\usepackage{amsthm}
\usepackage{amsfonts}
\usepackage{booktabs}
\usepackage{algorithm}
\usepackage{algorithmic}
\newcommand{\citeay}[1]{\citeauthor{#1}\,\shortcite{#1}}

\title{Reducing Bus Bunching with Asynchronous Multi-Agent Reinforcement Learning}

\author{
Jiawei Wang
\And
Lijun Sun\footnote{Contact Author}
\affiliations
McGill University, Montreal, Canada
\emails
jiawei.wang4@mail.mcgill.ca,
lijun.sun@mcgill.ca
}

\begin{document}

\maketitle

\begin{abstract}
The bus system is a critical component of sustainable urban transportation. However, due to the significant uncertainties in passenger demand and traffic conditions, bus operation is unstable in nature and bus bunching has become a common phenomenon that undermines the reliability and efficiency of bus services. Despite recent advances in multi-agent reinforcement learning (MARL) on traffic control, little research has focused on bus fleet control due to the tricky asynchronous characteristic---control actions only happen when a bus arrives at a bus stop and thus agents do not act simultaneously. In this study, we formulate route-level bus fleet control as an asynchronous multi-agent reinforcement learning (ASMR) problem and extend the classical actor-critic architecture to handle the asynchronous issue. Specifically, we design a novel critic network to effectively approximate the marginal contribution for other agents, in which graph attention neural network is used to conduct inductive learning for policy evaluation. The critic structure also helps the ego agent optimize its policy more efficiently. We evaluate the proposed framework on real-world bus services and actual passenger demand derived from smart card data. Our results show that the proposed model outperforms both traditional headway-based control methods and existing MARL methods.
\end{abstract}

\section{Introduction}

\begin{figure}[!b]
\centering
 \includegraphics[scale=0.3]{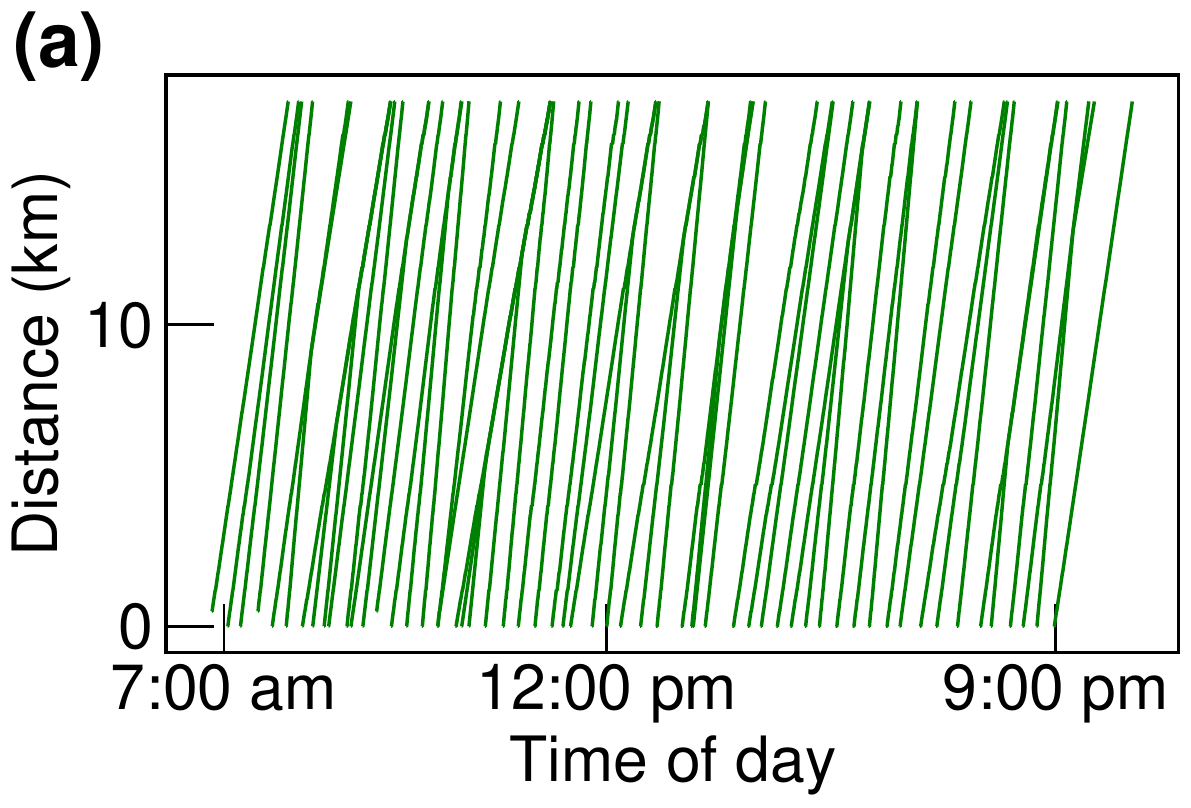} \qquad
\includegraphics[scale=0.3]{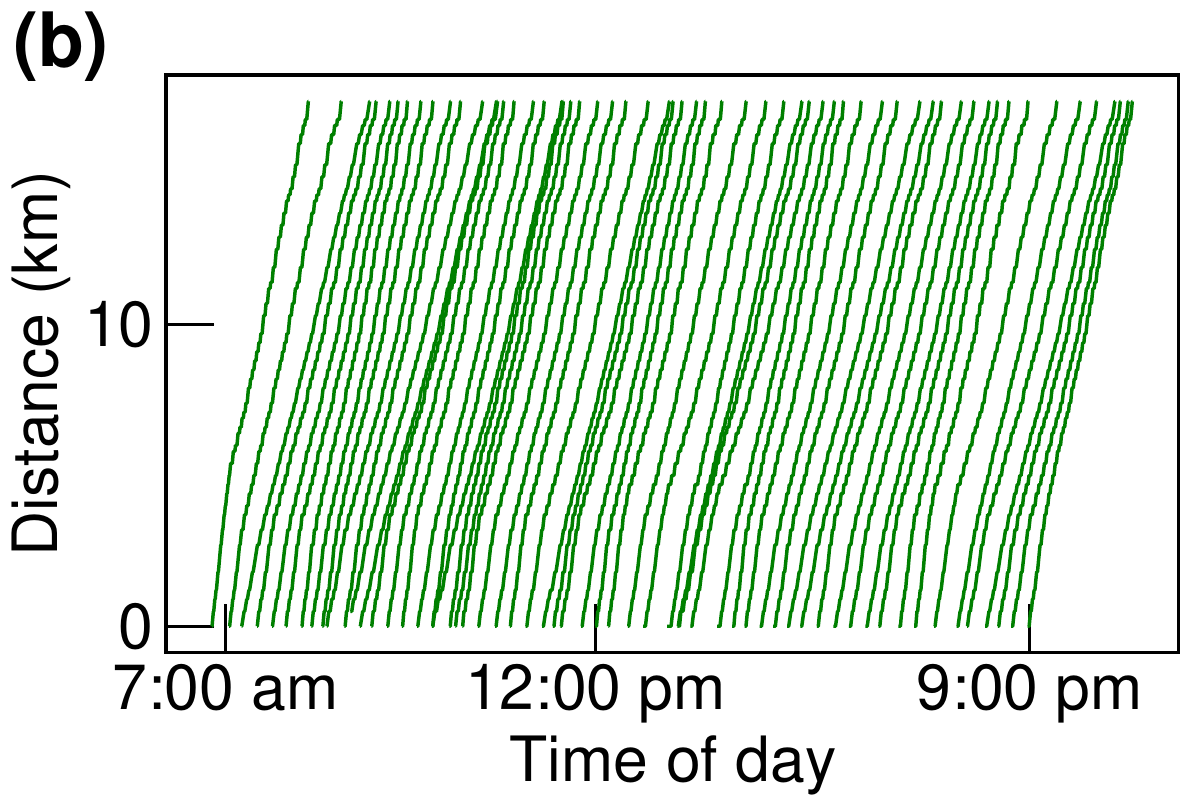}
\caption{Visualization of bus trajectories with heavy/reduced bus bunching in one day. Each line shows the trajectory of a bus. Panel (a) shows the trajectories over a day without any control/interventions, and panel (b) shows the results of our proposed CAAC framework. The negative effects of bus bunching are more severe on long truck services than on short feeder services.}
\label{bb}
 \end{figure}

Public transport has been considered the most critical component in sustainable urban transportation. Providing reliable and efficient services to meet the increasing transit demand has become a major challenge faced by public transit operators. However, in real-world operations, the great uncertainties in traffic conditions (e.g., congestion, incident, and signal control) and dwell time often make bus services unstable \cite{daganzo2019public}. Bus bunching has become a common phenomenon in the operation of high-frequency bus services. For a bus service, vehicles essentially leave the departure terminal with a regular frequency/headway (e.g., every 10 min) based on the timetable. If a bus is slightly delayed on the route, it will likely encounter more waiting passengers at the next bus stop, and then the dwell time serving boarding/alighting passengers will also increase. As a result, the delayed bus will be further delayed, and the following bus will become faster due to fewer waiting passengers and shorter dwell time; eventually, two or even more buses will bunch together and travel as a group (e.g., see Fig.~\ref{bb}). Bus bunching has two major negative effects on rider experience: on the one hand, passengers will suffer from long waiting times and then see a group of vehicles arriving together; on the other hand, bus bunching also leads to imbalanced occupancy, since the leading bus in a group will take more passengers and the others will take less. Overall, bus bunching is an urgent issue that undermines the reliability and efficiency of transit services. In general, the severity of bus bunching increases with the service length, and downstream passengers often suffer more than upstream passengers.

Bus holding control is one of the most effective methods to reduce bus bunching. Recently advances in model-free reinforcement learning (RL) have shed new light on the sequential control problem, allowing us to develop efficient and effective control policies for a large system without building an explicit model. In particular, multi-agent reinforcement learning (MARL) provides a learning scheme to explore collaborative policy, which has been shown to be superior in a multi-agent system. The application of MARL in intelligent transportation systems (ITS) has attracted considerable attention in recent years, with a particular focus on traffic signal control (see e.g., \cite{chen2020toward}). Bus fleet operation is also a unique ITS application where MARL fits very well. On this track, \citeay{chen2016real} and \citeay{alesiani2018reinforcement} proposed MARL models to develop holding control strategies to match a pre-specified headway; however, such a setting is often too restrictive to adapt to the great uncertainties in bus fleet operation. Instead of using a fixed headway, \citeay{wang2020dynamic} proposed a reward function to promote headway equalization in the MARL model. Nevertheless, the models above overlook the fact that buses are not being controlled simultaneously since holding happens only when a bus arrives at a bus stop. In such an asynchronous setting, the default MARL cannot effectively take into account the impact from other agents. To deal with such asynchronicity, \citeay{menda2018deep} proposed an event-driven MARL by designing a reward function associated with macro-action; however, this approach still cannot distinguish the contributions from ego agent and other agents.

To better address the asynchronous issue in bus fleet control, in this paper we propose a new credit assignment framework for asynchronous control (CAAC) by integrating actions from other agents using inductive graph learning. Our main contribution is threefold:
\begin{itemize}
\item We establish an asynchronous multi-agent reinforcement learning (ASMR) framework to optimize holding control policy on a bus route to reduce bus bunching.
\item To the best of our knowledge, this work is the first to consider the marginal contribution from uncertain events in MARL. This proposed CAAC also has potential to be generalized to other asynchronous control problems.
\item We build a learning framework for bus operation in an asynchronous setting based on real-world transit operation data, and evaluate the proposed CAAC with extensive experiments.
\end{itemize}

\section{Related Work}

\subsection{Holding Control to Avoid Bus Bunching}
Bus holding control is a longstanding research topic in transit operation \cite{daganzo2019public}. Traditional methods mainly focus on maintaining headway consistency by deriving suitable policies using optimization-based models. However, in practice, this approach has two major limitations. On the one hand, due to the lack of field data, scenarios and optimization models are often over-simplified in many existing studies (e.g.,  \cite{daganzo2009headway,wang2020dynamic}). As a result, these models can hardly reproduce reality and scale to large real-world scenarios. On the other hand, these optimization-based models are essentially scenario-specific \cite{seman2019headway}, and in practice they require careful design and training for each case. Therefore, these optimization-based models often fail to generalize the policy learned from the past to future and unseen scenarios.

\subsection{Asynchronous Multi-agent Control}
In this paper, we tackle the asynchronous multi-agent control problem by introducing a new credit assignment scheme. Credit assignment aims to better estimate each agents' own contribution to the team's success for better policy update, and it has been a central research topic in designing reliable MARL algorithms \cite{chang2004all}. For example, \citeay{foerster2018counterfactual} proposed COMA---a multi-agent actor-critic algorithm with a counterfactual baseline to quantify the contribution from each agent. However, COMA only works for applications with a discrete action space. Recent studies have proposed credit assignments by designing different value decomposition schemes, which can accommodate both discrete and continuous action spaces. For example, \citeay{sunehag2018value} proposed Value-Decomposition Network (VDN) to measure the impact of each agent on the observed joint reward. \citeay{rashid2018qmix} developed QMIX, which improves VDN by adding restrictions on the relation between local and global Q-value. Furthermore, \citeay{son2019qtran} developed QTran as a more generalized factorization scheme. These studies have made outstanding contributions to the general credit assignment problem. However, a major limitation is that they only consider local observation-action, which hinders the exploration of more effective cooperation policy. Notably, \citeay{sqv} derived a similar credit assignment framework with Shapely Q-value to better quantify local contributions; however, the asynchronous issue with a varying number of activated agents is still overlooked. \citeay{lee2020reinforcement} studied sequential decision problems involving multiple action variables whose control frequencies are different; still, the framework only considers a pre-specified frequency for each single agent.


\section{Holding Control as an ASMR Task}

In this paper, we define the forward headway for a specific bus at a certain time as the time duration for it to reach the current location of the preceding bus. Correspondingly, we define the backward headway as the forward headway of its follower. Holding control is a widely used strategy to maintain headway consistency and avoid bus bunching in daily operation \cite{wang2020dynamic}. The key idea of holding control is to let a bus stay longer (i.e., by adding a slack period) at stops in addition to the dwell time for passengers to board/alight. For instance, when a bus gets too close to the preceding vehicle, it can equalize/balance the forward headway and backward headway by holding at a bus stop. In this study, we model bus holding as a control policy in MARL with the goal of improving overall operation efficiency.

\begin{figure}[!t]
\centering
     \includegraphics[scale=0.25]{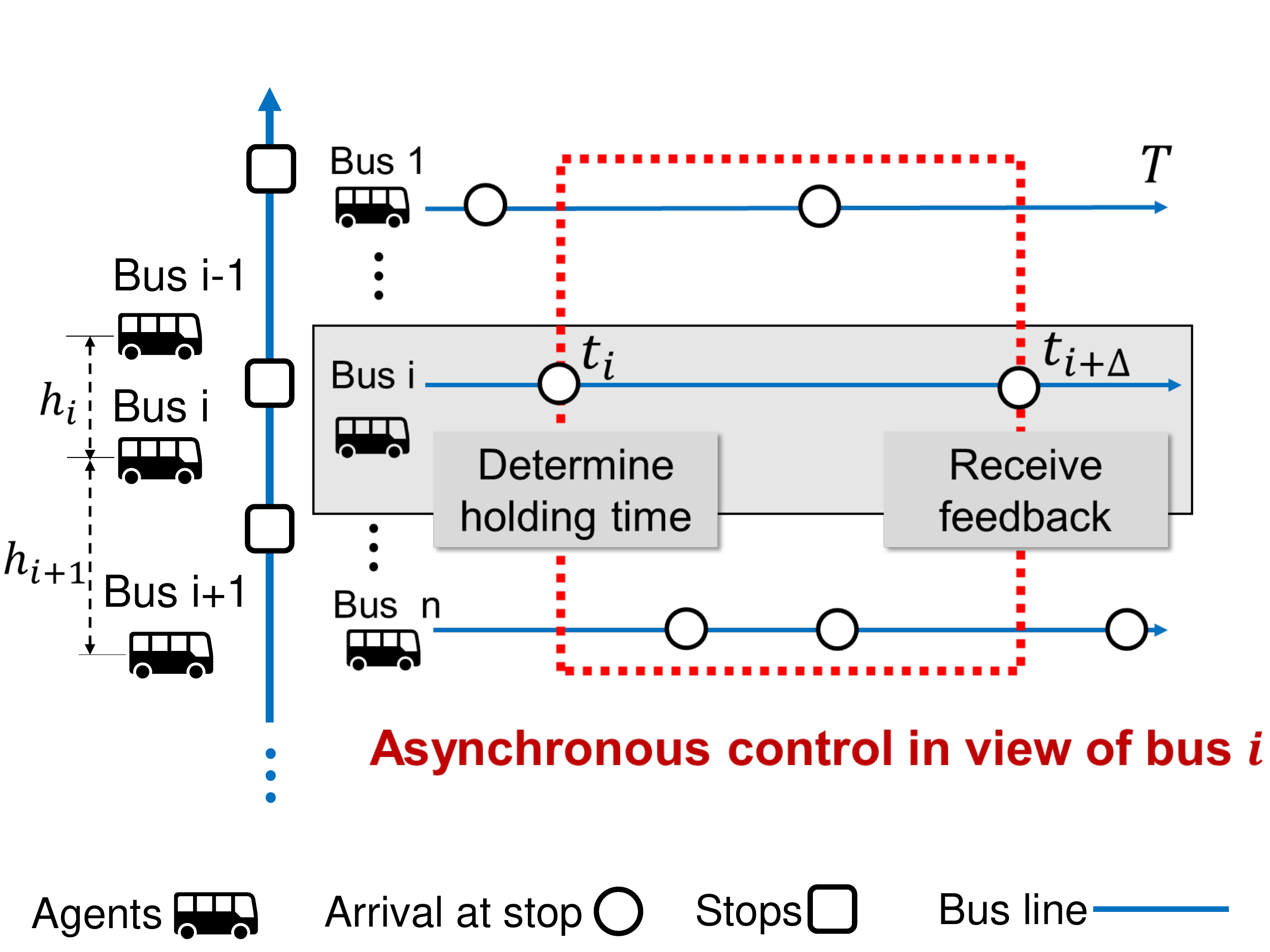}
  \caption{The vehicle control framework on a bus route. A circle indicates the event of a bus arriving at a bus stop. For bus $b_i$ at time $t_i$, we denote by $h_{i+1}$ the backward headway (time for the following bus $b_{i+1}$ to reach the current location of $b_i$) and $h_{i}$ the forward headway (time for $b_i$ to reach the current location of $b_{i-1}$).}
   \label{fig:framework}
\end{figure}

Fig.~\ref{fig:framework} shows the vehicle holding control framework on a bus route, in which we take all the buses running in the system/environment as agents. Note that we label all buses in a sequential order, where bus $b_{i+1}$ follows bus $b_i$. Bus holding control for a fleet $\{b_i\}$ is a typical application of asynchronous multi-agent control, as holding control is applied only when a bus arrives at a bus stop. We define the basic elements in this framework as follows:

\noindent\textbf{State:} Each agent has a local observation of the system. For bus $b_i$ arriving at stop $k_j$ at time $t$, we denote the state of $b_i$ by $s_{i,t}$, which includes the number of passengers onboard, the forward headway $h_i$ and the backward headway $h_{i+1}$.

\noindent\textbf{Action:} Following \citeay{wang2020dynamic}, we model holding time as:
\begin{equation}
\Delta d_{i,t}=a_{i,t} \Delta T,
\label{eq:action}
\end{equation}
where $\Delta T$ is the maximum holding duration and $a_{i,t} \in \left[0,1\right]$ is a strength parameter. We consider $a_{i,t}$ the action of bus $b_i$ when arriving at a bus stop at time $t$. Here, $\Delta T$ is used to limit the maximum holding duration and avoid over-intervention. We model $a_{i,t}$ as a continuous variable in order to explore the near-optimal policy in an adequate action space. Note that no holding control is implemented when $a_{i,t}=0$.

\noindent\textbf{Reward:} Despite that holding control can reduce the variance of bus headway and  promote system stability, the slack/holding time will also impose additional penalties on both passengers (i.e., increasing travel time) and operators (i.e., increasing service operation hours). To balance system stability and operation efficiency, we design the reward function associated with bus $b_i$ at time $t$ as:
\begin{equation}
{r_i^t}= -(1-w)\times CV^2-w\times a_{i,t},
\label{eq:reward}
\end{equation}
where $CV^2=\frac{Var[h]}{E^2[h]}$ quantifies headway variability \cite{NAP24766} and $w\in \left[0,1\right]$ is a weight parameter. Essentially, $E[h]$ is almost a constant given the schedule/timetable, and thus $CV$ is mainly determined by $Var[h]$: a small $CV$ indicates consistent headway values on the bus route, and a large $CV$ suggests heavy bus bunching. The second term in Eq.~\eqref{eq:reward} penalizes holding duration and prevents the learning algorithm from making excessive control decisions. This term is introduced because any holding decisions will introduce additional slack time and reduce operation efficiency. Overall, the goal of this reward function is to effectively achieve system stability with as few interventions as possible, with $w$ serving as a balancing parameter. 

The agent implements an action (i.e., determining holding time) when it arrives at a stop, and then it will receive feedback upon arriving at the next stop. In previous studies \cite{wang2020dynamic,chen2016real}, this dynamic process is considered a multi-agent extension of Markov decision processes (MDPs) \cite{littman1994markov} and referred to as Markov game $G=\left(N,\mathcal{S}, \mathcal{A},\mathcal{P}, \mathcal{R},{\gamma}\right)$, where $N$ denotes the number of agents, $\gamma$ denotes the discount factor, $\mathcal{S}$ and $\mathcal{A}=\left\{A_1,\ldots,A_N\right\}$ denote the state space and the joint action space, respectively, and $\mathcal{P}:\mathcal{S}\times\mathcal{A}\mapsto\mathcal{S}$ represents the state transition function. However, this definition becomes inaccurate in the bus control problem, since agents rarely implement actions at the same time. To address this asynchronous issue, we introduce a modified formulation in which each agent maintains its own transition: $\hat{\mathcal{P}_i}:\mathcal{S}_i\times\hat{\mathcal{A} }\mapsto\mathcal{S}_i$, $ \hat{\mathcal{A}} \subseteq \mathcal{A}$, where $S_i$ is the observation of agent $i$. In this way, state transition observed by agent $i$ does not necessarily depend on the actions from all the other agents at any particular time. The policy to choose an action is given by the conditional probability $\pi_{\theta_i}=p\left(A_i\mid S_i\right)$. Finally, the reward function for each agent can also be defined independently: $\mathcal{R}=\left\{R_1, \ldots,R_N\right\}$. While the independence assumption simplifies the problem, it raises a new challenge---how to effectively consider the actions from other agents, which is to be addressed in the following section.

\section{Methods}

As mentioned previously, traditional multi-agent models, such as Independent Q-Learning (IQL) \cite{tan1993multi}, MADDPG \cite{lowe2017multi} and QTran \cite{son2019qtran}, cannot address the aforementioned asynchronous issue. To solve this problem, in this section we introduce a Credit Assignment framework for Asynchronous Control (CAAC) with shared parameters.

\subsection{Credit Assignment Framework for Asynchronous Control (CAAC)}
\begin{figure}[!b]
\centering
     \includegraphics[scale=0.2]{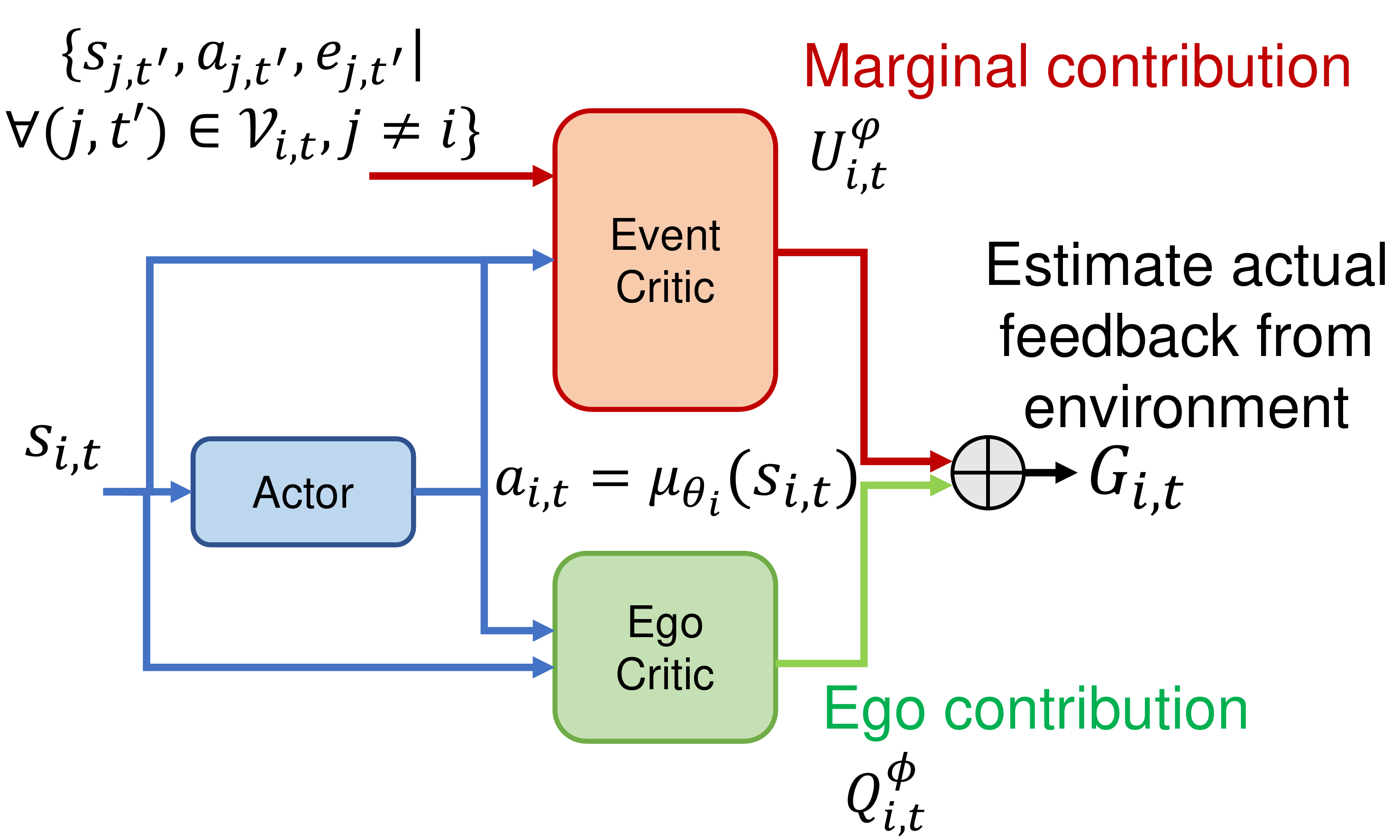}
  \caption{Architecture of CAAC.}
   \label{fig:CAAC}
\end{figure}

Following previous work, we adopt the basic actor-critic architecture, in which the critic is used to evaluate the effect of action from a single agent. To achieve reliable and efficient credit assignment in an asynchronous setting, we design an inductive critic following the scheme shown in Fig.~\ref{fig:CAAC}.

The proposed inductive critic consists of an ego critic and an event critic: the ego critic is to evaluate the actor's policy, and the event critic is to account for the contribution from the actions of other agents. With this architecture, we expect CAAC to better quantify the undetermined impact from other agents and provide a more accurate credit assignment scheme for policy evaluation. We next introduce CAAC and the learning algorithm in detail.

\subsubsection{Approximate Marginal Contribution with Event Critic}
\begin{figure}[!t]
\centering
\includegraphics[scale=0.2]{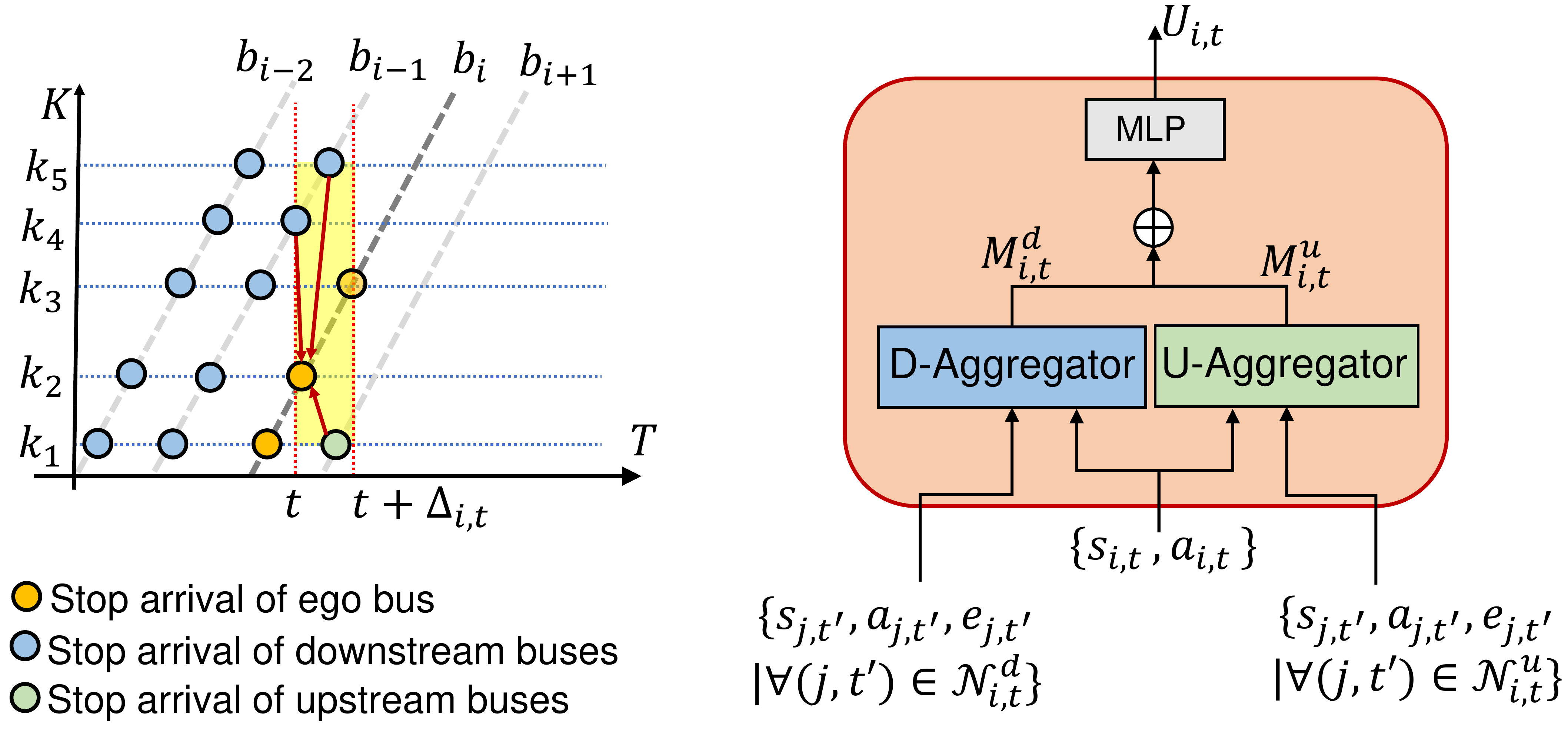}
  \caption{{Event Graph} (EG) at time $t+\Delta_{i,t}$ (left) and the corresponding {Event Critic} (right) for agent/bus $b_i$. The dashed lines in EG show bus trajectories. We highlight the three incoming edges (in red) for event $\left(i,t\right)$. D-aggregator and U-aggregator indicate aggregators for downstream events and upstream events, respectively.}
   \label{fig:EC}
\end{figure}

Inspired by the idea of inductive graph learning \cite{hamilton2017inductive}, we propose to build a graph-based event critic to learn to approximate the marginal contribution from other agents. We first introduce the concept of event graph (EG) based on the spatiotemporal trajectory plot (see Fig.~\ref{fig:EC}). We denote EG by ${G}\left({V},{E}\right)$, where ${V}$ and ${E}$ are the vertex set and edge set, respectively. We define a vertex $(i,t)$ as the event when a bus $b_i$ arrives at a bus stop at time $t$. Note that we ignore the index of bus stops when defining EG. We assume bus $b_i$ arrives at the next stop at time $t+\Delta_{i,t}$, and introduce incoming edges for $(i,t)$ by linking all arriving events $\left(j,t'\right)$ from other vehicles (i.e., $j\neq i$) within the time window $t<t'\le t+\Delta_{i,t}$. The left panel of Fig.~\ref{fig:EC} illustrates how EG is built based on the trajectory diagram. Each node in the panel represents an arrival event of a bus. Taking the event $\left(i,t\right)$ as an example, there are three incoming edges from other events. We denote the set of neighbors of event $\left(i,t\right)$ by $\mathcal{N}_{i,t}=\left\{\left(j,t'\right) \in {V} : j\neq i, t<t'\le t+\Delta_{i,t} \right\}$.


Given an ego agent, the formulation of EG presents a neat and dynamic tool to account for actions from other agents, which allows us to conduct inductive graph learning and build a flexible policy evaluation block for asynchronous control. Specifically, we adopt graph attention neural network (GAT) \cite{velivckovic2017graph} to inductively learn and approximate the undetermined impact from the actions of other agents. Considering the heterogeneous effects from upstream and downstream events, we introduce two separate GATs for upstream events and downstream events, respectively. In doing so, we divide $\mathcal{N}_{i,t}$ into two subsets: upstream event set $\mathcal{N}^u_{i,t}=\left\{\left(j,t'\right) \in {V} : j>i, t<t'\le t+\Delta_{i,t} \right\}$  and downstream event set $\mathcal{N}^d_{i,t}=\left\{\left(j,t'\right) \in {V} : j< i, t<t'\le t+\Delta_{i,t} \right\}$. The right panel of Fig.~\ref{fig:EC} summarizes the inductive block, which we call event critic (EC).


We next introduce the details of EC. We first define $e_{j,t'}=\left\{e^1_{j,t'},e^2_{j,t'}\right\}, \forall  \left(j,t'\right)\in \mathcal{N}_{i,t} $ to capture the augmented information for the event critic, where $e^1_{j,t'}$ denotes the number of bus stops separating the two controls and it is normalized by the total number of stops in the system, and $e^2_{j,t'}=|j-i|$ is the number of vehicles in between $b_i$ and $b_j$. We then aggregate information from upstream and downstream, respectively. 
Given the upstream events $\mathcal{N}^u_{i,t}$, the node features consist of two parts: (1) node features of other agents:  $h_{j,t'} = \left(s_{j,t'},a_{j,t'},e_{j,t'} \right)$, where $\left(s_{j,t'},a_{j,t'}\right)$ is the state-action pair of agent $j$ at $t'$; (2) node feature of the ego agent $h_{i,t}=\left(s_{i,t},a_{i,t},\text{zeros\_like}\left(e_{j,t'}\right) \right)$, where we add zeros to ensure equal feature length. Since EG is undirected, the graph attention is actually based on the relation between the ego event and its upstream events. We can then perform graph attention:
\begin{gather}
 \alpha_{(i,t),(j,t')}=\frac{\exp\left(\sigma\left(f\left(W^ah_{i,t}|| W^ah_{j,t'}\right)\right)\right)}{\sum_{\left(j,t'\right)\in \mathcal{N}^u_{i,t}}\exp\left(\sigma\left(f\left(W^ah_{i,t}||W^ah_{j,t'}\right)\right)\right) }
\label{eq:att}\\
 h'_{i,t} = \sum\nolimits_{\left(j,t'\right)\in \mathcal{N}^u_{i,t}} \alpha_{(i,t),(j,t')} W^ah_{j,t'},
\label{eq:ego}
\end{gather}
where $\sigma$ is non-linear activation function and $||$ represents concatenation, $W^a$ is the trainable shared weight matrix, and $f$ represents a single-layer neural network. Finally, we sum all the derived features as
\begin{equation}
 M^u_{i,t} = \sum\nolimits_{k \in \left\{ \left(i,t\right) \cup \mathcal{N}^u_{i,t}\right\}} \sigma \left( h_{k} \right).
\label{eq:sum}
\end{equation}

The information aggregation $M^d_i$ for downstream events $\mathcal{N}^d_{i,t}$ works the same way as the upstream.
The summation of $M^u_i$ and $M^d_i$ is then fed to a feed-forward neural network to generate the final result of the event critic. Note that we omit self-attention and mask the aggregation output if there exist no upstream or downstream events for the ego agent (i.e., no contribution from other agents).

\subsubsection{Deep Deterministic Policy Gradient with Inductive Critic}

A critical issue in MARL with Deep Deterministic Policy Gradient (DDPG) is that the critic is likely to have an ambiguous evaluation on the contribution of each agent. The problem becomes even worse in an asynchronous setting where even the number of activated agents varies in each decision step. To address this issue, we implement multi-agent DDPG with a more rational credit assignment by combining the ego critic and event critic (see Fig.~\ref{fig:CAAC}).


As the first component of the inductive critic, the ego critic is responsible for evaluating the ego policy. Based on the evaluation, we can derive policy gradient to maximize the following objective function as in DDPG:
\begin{equation}
\mathcal{J}(\theta ) = \mathbb{E} \left[ Q^{\phi }_{i,t}\left(s_{i,t},\mu_{\theta }\left(s_{i,t}\right)\right) \right].
\label{eq:actor}
\end{equation}
The second part of the inductive critic is the event critic, which is used to approximate the marginal contribution from other agents. Finally, we approximate the actual system return for agent $i$ using the sum of the output from ego critic and event critic:
\begin{equation}
G_{i,t} = Q^{\phi}_{i,t} +U^{\psi}_{i,t }.
\label{eq:sumqu}
\end{equation}

Based on the standard DQN loss \cite{mnih2015human} following the  Bellman equation, the proposed inductive critic can be trained by minimizing the following loss function:
\begin{equation}
\begin{split}
&\mathcal{L}(\phi ,\psi ) = \mathbb{E}  \bigg[\left(r_{i,t} +\gamma G^{\text{tar}}_{i,t+\Delta_{i,t}}  - G_{i,t}  \right)^2  \\
&+\beta \left(
\mathbb{I}({{N}^u_{i,t}  =\emptyset })
\left\| M^u_{i,t} \right\|_2 + \mathbb{I}({{N}^d_{i,t}  =\emptyset })   \left\| M^d_{i,t} \right\|_2\right)\bigg],
\label{eq:critic}
\end{split}
\end{equation}
where $r_{i,t}$ is the reward obtained by agent $i$ given its action $a_{i,t}$, $G^{\text{tar}}_{i,t+\Delta_{i,t}}$ denotes the cumulative return from the next decision step $t+\Delta_{i,t}$ in the view of agent $i$, which is estimated through the target network \cite{mnih2015human}, and $\mathbb{I}(\cdot)$ is an indicator function.
Note that in the case where an event has no upstream/downstream events, we additionally impose L2 norm with hyperparameter $\beta=0.1$. Thus, $\left\| M^u_{i,t} \right\|_2$ and $\left\| M^d_{i,t} \right\|_2$ regularize $W^a$ in the event critic block, which can facilitate inductive learning. In training, we sample from the interaction between agents and the system to approximate the expectation of above objective functions. In the execution phase, only local observations are used for actor network; thus, the overall CAAC is very flexible for real-world applications.

\section{Evaluation}

We evaluate the performance of the proposed method based on real-world data. Four bus routes (R1-R4) in an anonymous city are selected, with true passenger demand derived from smart card data. The four routes are all trunk services covering more than 15 km with over 40 bus stops along the route. Table~\ref{lines} lists the basic statistics of the routes, including the number of services per day, the number of stops, route length, the mean and standard deviation (std) of headway at the departure terminal.

\begin{table}[!ht]
	\centering
    \footnotesize
\begin{tabular}{ccccccc}
\toprule
		 & services  & stops & length (km) & mean (sec) & std (sec)  \\
		\midrule
		R1 & 59  &46 & 17.4 &874 & 302  \\
		R2 & 72 &58 & 23.7 &745 & 307  \\
		R3 & 57  &61  & 23.2 &931 &354 \\
		R4   & 55  &46 & 22.5 &955 &351 \\
		\bottomrule
	\end{tabular}
    \caption{Basic information of bus lines.}
    	\label{lines}
\end{table}

\subsection{Simulator for Bus Route Operation}
\begin{figure}[!b]
\centering
 \includegraphics[scale=0.3]{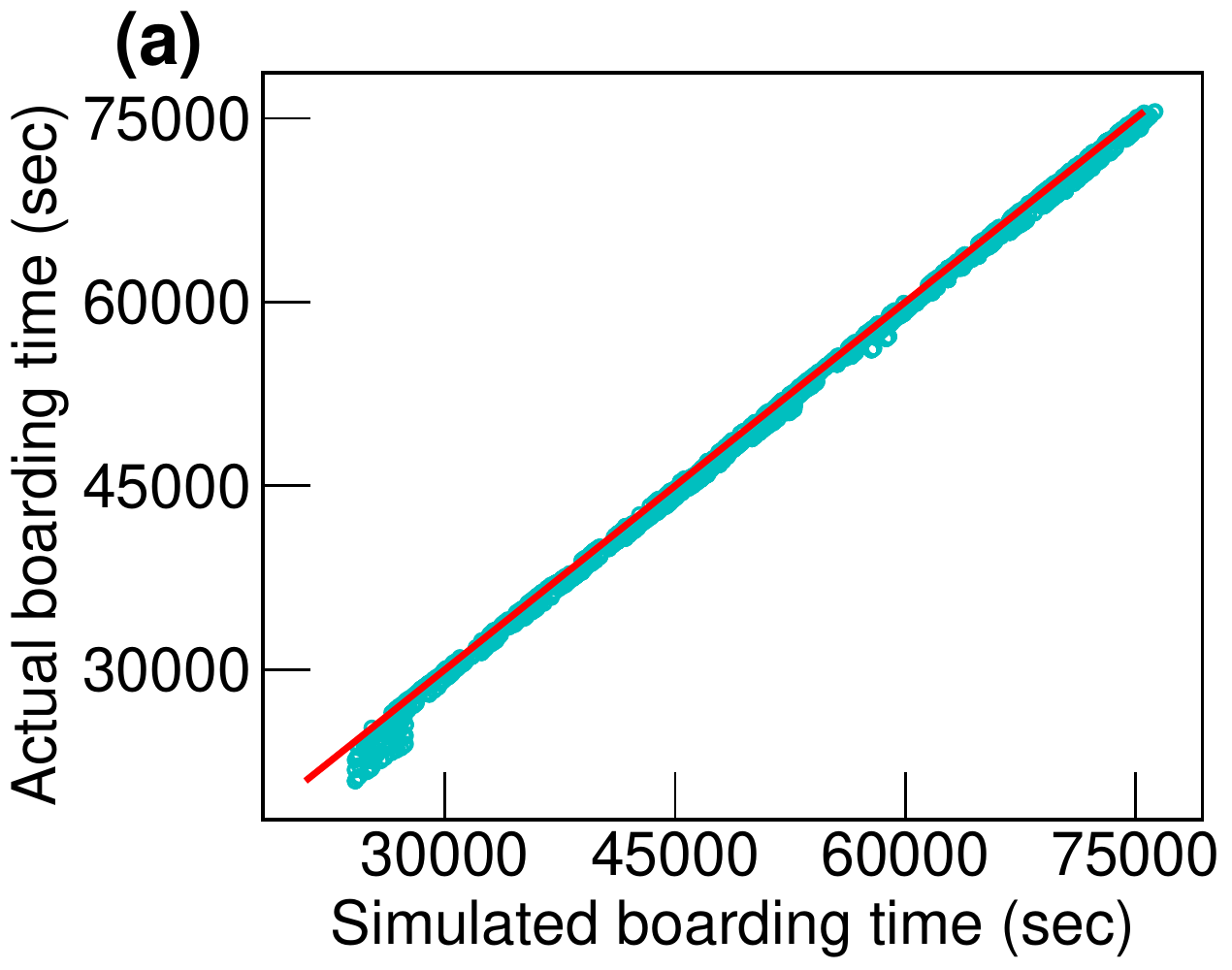} \qquad
\includegraphics[scale=0.3]{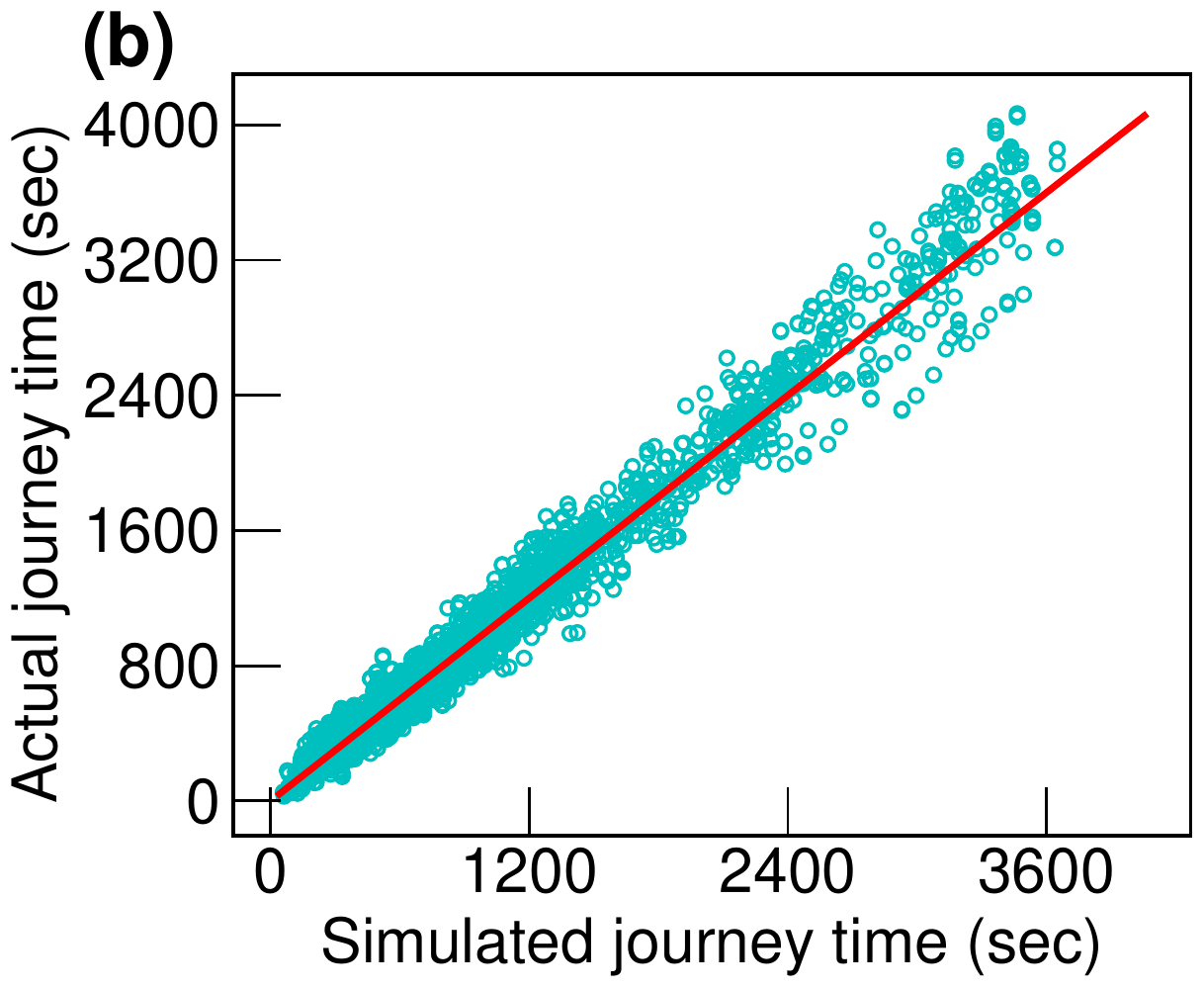}
\caption{Simulated values v.s. true values from smart card data on R1: (a) boarding time (i.e., smart card tapping-in time), and (b) journey time (i.e., duration between tapping-in and tapping-out). }
   \label{sim}
 \end{figure}
We first develop and calibrate the bus simulator to reproduce the patterns of real-world operation.
In this simulation, the alighting and boarding times per passenger are set to $t_a=1.8\ \text{s/pax}$ and $t_b=3.0 \ \text{s/pax}$, respectively. To simulate the uncertainty of road conditions, buses are given a random speed $v\times\mathcal{U}(0.6,1.2)$ km/h when travelling between every two consecutive stops, where $v$ is set to 30 km/h and $\mathcal{U}$ denotes a continuous uniform distribution. The capacity of the bus is set to 120 $\text{pax}$.
Fig.~\ref{sim} (a) and (b) show the simulated boarding time and actual boarding time from smart card (tap-in for boarding and tap-out for alighting) data and the simulated journey time and actual journey time, respectively, for service R1. The Pearson correlations for these two plots are 0.999 and 0.983, respectively, suggesting that our simulator is consistent with the real-world operation.

\subsection{Experimental Settings}
We set hyper-parameter $w=0.2$ in reward function Eq.~\eqref{eq:reward} to place priority on system stability. We start our experiment by training the model on R1 for 250 episodes. In the execution phase, we test the performance of the tuned RL models on R1 and then evaluate the model transferability by applying R1's models on R2/R3/R4 directly without retraining.
The proposed \textbf{CAAC} framework is compared with the following baseline models, including both traditional headway-based control models and state-of-the-art MARL methods:
i) \textbf{No control (NC)}: NC is considered a naive baseline where no holding control is implemented.
ii) \textbf{Forward headway-based holding control (FH) } \cite{daganzo2009headway}: the holding time is $d=\max \{ 0, \overline{d}+g(H_0-h^-) \}$, where $H_0$ is the desired departure headway, $h^-$ is the  forward headway of the arriving bus, $\overline{d}$ is the average delay at equilibrium, and $g>0$ is a control parameter. We use the same parameters as in \citeay{daganzo2009headway}.
iii) \textbf{Independent Actor-Critic (IAC)} \cite{lillicrap2015continuous}: we implement DDPG in the IAC setting to examine the performance where the agent completely overlooks the impact from other agents. IAC can be considered a special case of CAAC with no event critic. iv) \textbf{MADDPG} \cite{lowe2017multi}: we implement MADDPG as a state-of-the-art baseline without special consideration for the asynchronous setting. In MADDPG, the ego bus considers all other buses when performing actions. In the implementation, we fill zeros for the actions from inactivated agents. Besides, to achieve transferability, both the critic network and the actor network are shared among all agents.

All models are implemented with python and PyTorch 1.7.0 on Ubuntu 18.04 LTS, and experiments are conducted on a server with 256GB RAM.
We use the following indicators to evaluate model performance: i) \textbf{Average holding time (AHT)}, which characterizes the degree of intervention; ii) \textbf{Average waiting time (AWT)}, which evaluates the severity of bus bunching; iii) \textbf{Average journey time (AJT)}, which quantifies the average journey duration from boarding to alighting for all trips; iv) \textbf{Average travel time (ATT)}, which quantifies the average travel time for each bus from the departure terminal; and v) \textbf{Average occupancy dispersion (AOD)}, which  evaluates how balanced the occupancy is. Note the dispersion is computed as a variance-to-mean ratio of the number of onboard passengers. We expect to see a large AOD value when bus bunching happens, since the number of onboard passengers will be highly imbalanced in those situations.

\subsection{Results}

\subsubsection{Execution Results on Trained Route (R1)}

\begin{figure}[!t]
\centering
 {\includegraphics[scale=0.30 ]{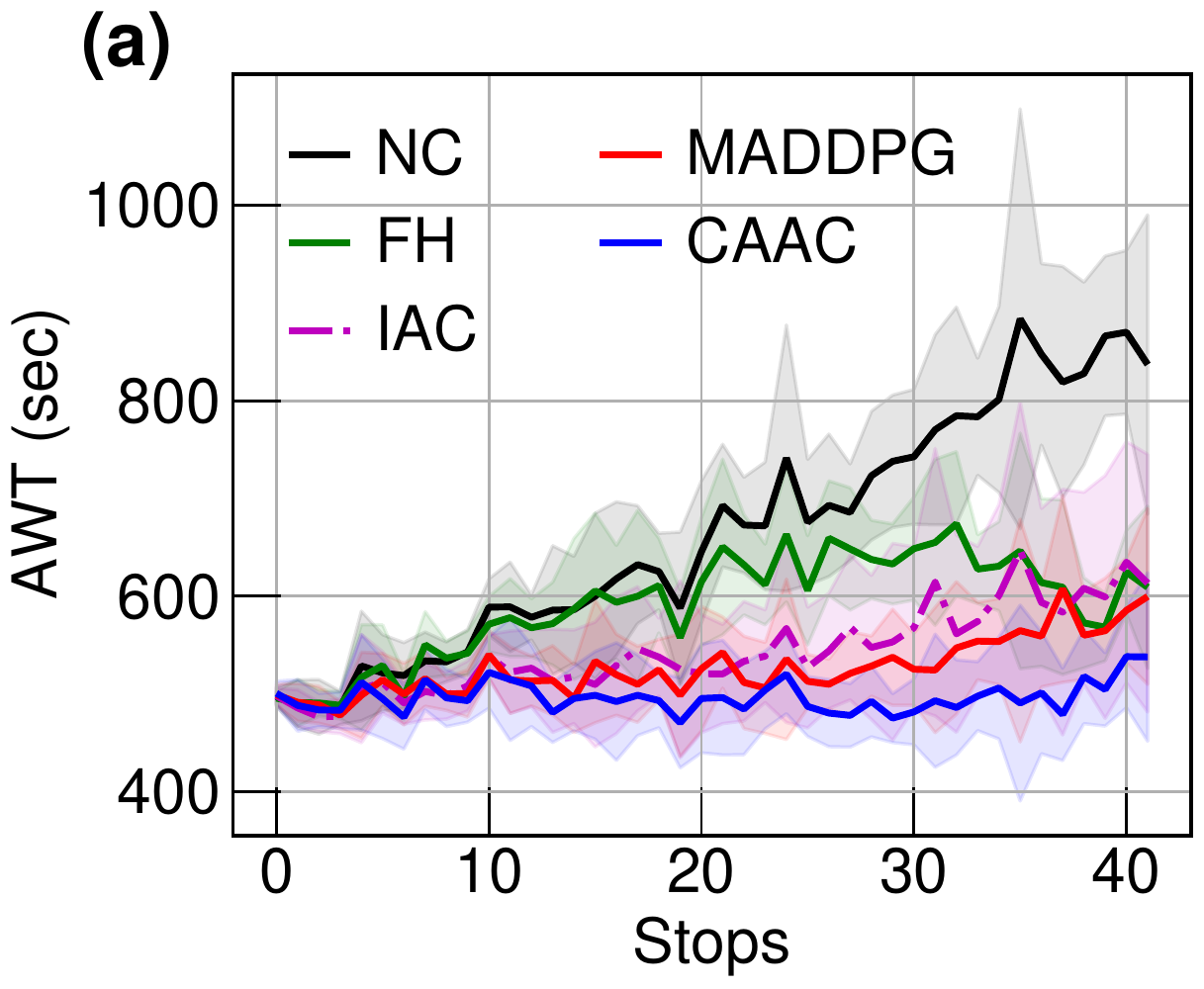}\label{fig:wait_w}}\qquad
 {\includegraphics[scale=0.30 ]{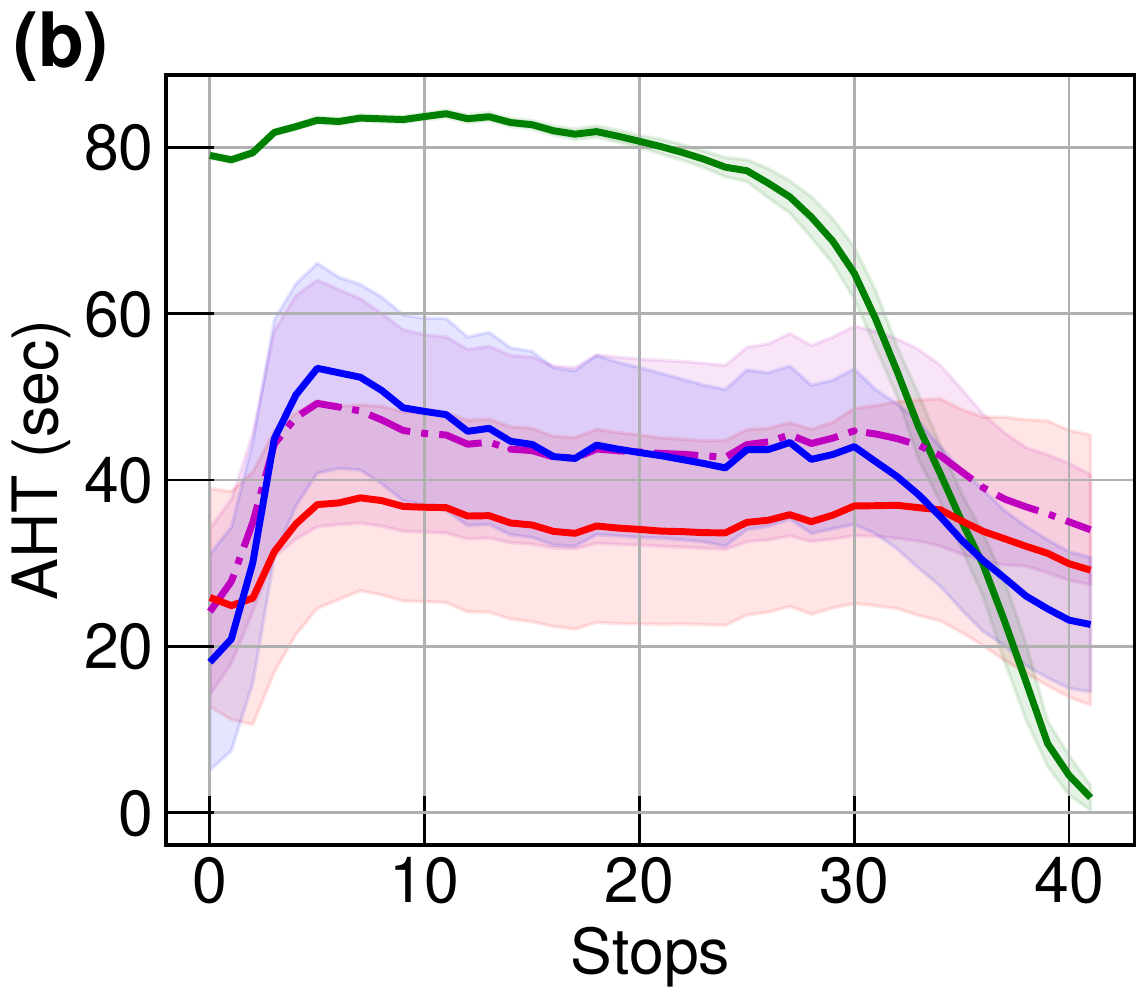}\label{fig:hold_w}}\\
 {\includegraphics[scale=0.30 ]{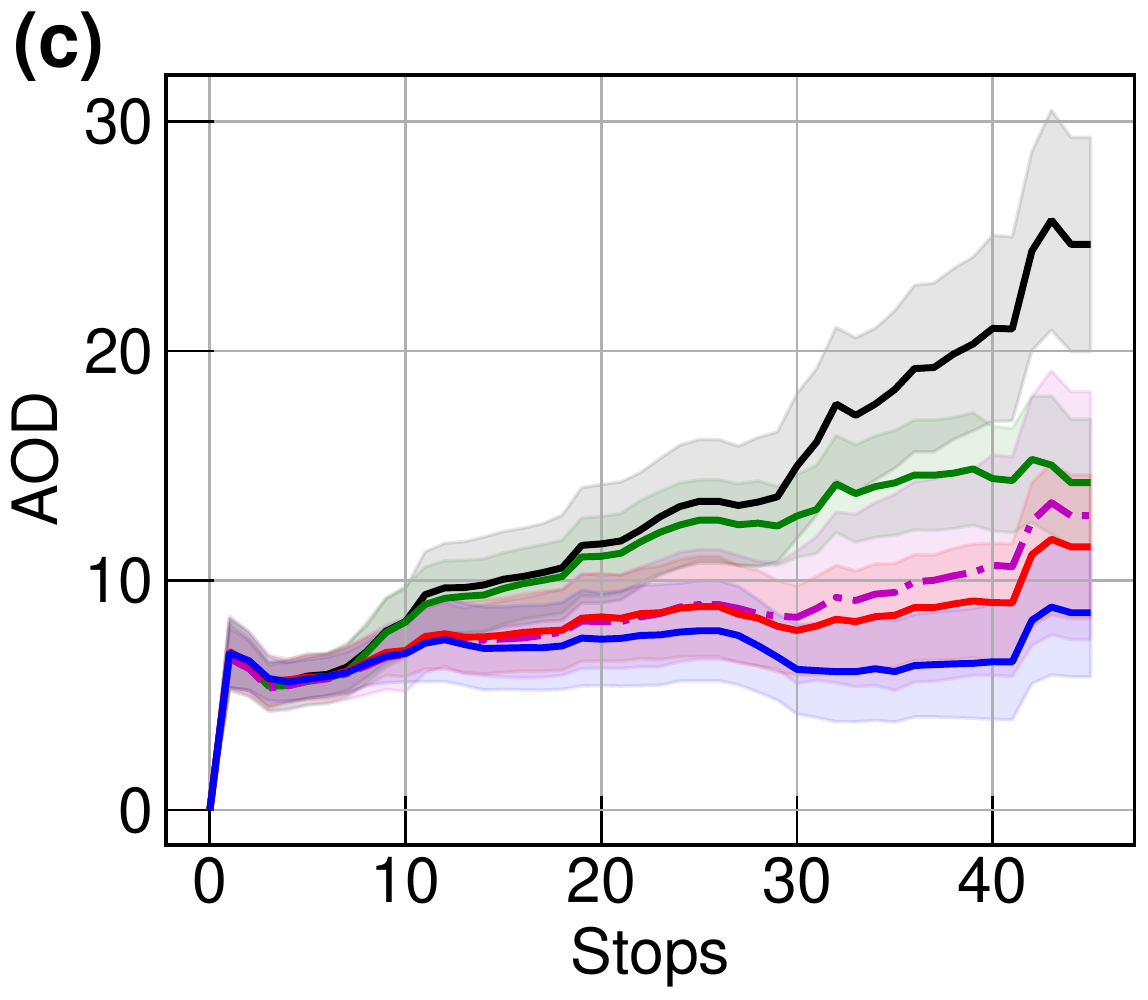}\label{fig:adt_w}}\qquad
 {\includegraphics[scale=0.30 ]{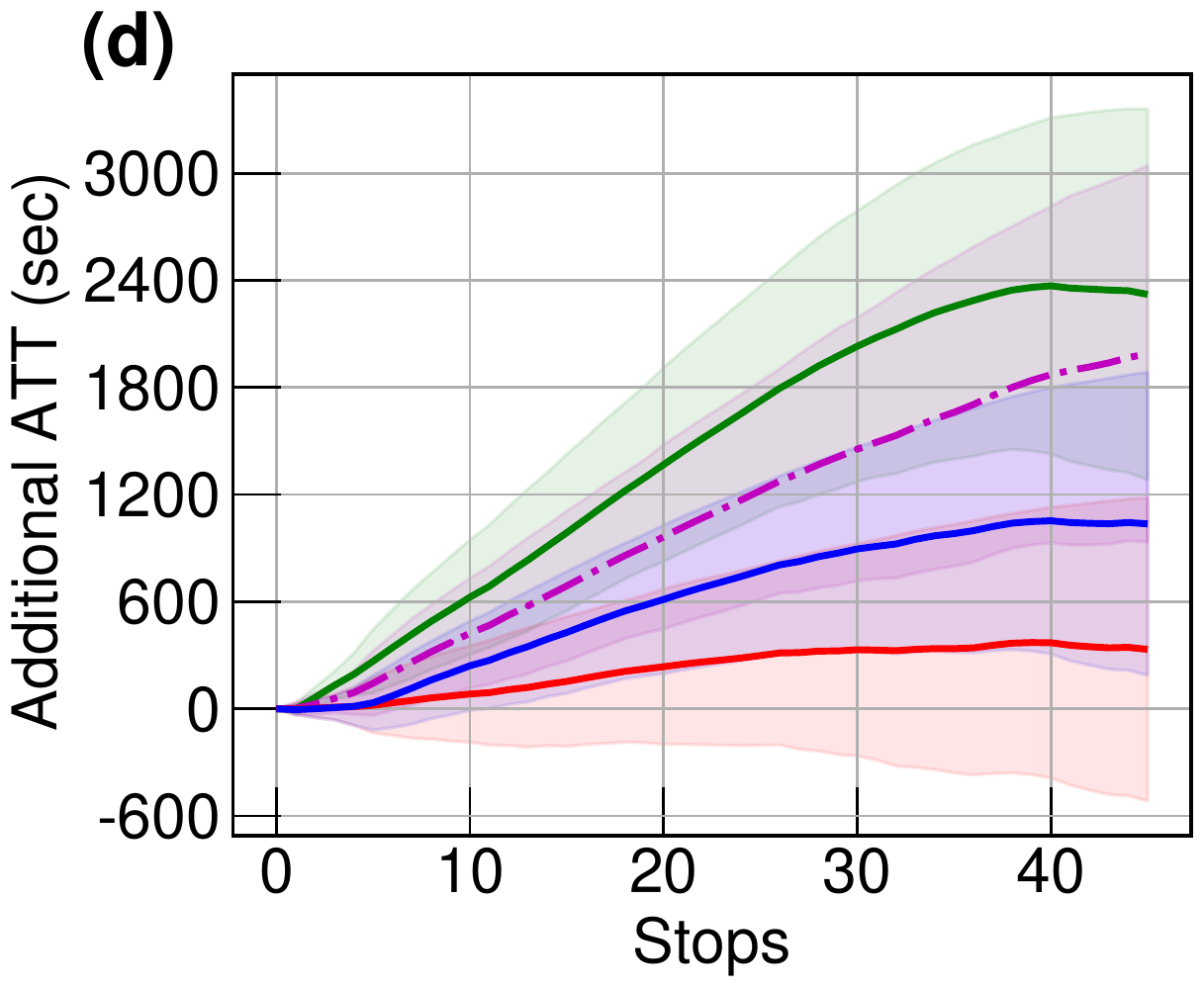}\label{fig:h_w}}
\caption{Stop-wise performance comparison: (a) average waiting time; (b) average holding time; (c) average occupancy dispersion; (d) additional travel time from departure terminal w.r.t. NC.}
\label{fig:sw}
\end{figure}

We first evaluate model performance on R1 on which all the MARL models are trained. Fig.~\ref{fig:sw} presents detailed stop-wise performance comparison. Overall, the result suggests that RL-based models perform better in reducing bus bunching at a lower cost than traditional headway-based control FH. We believe it is because of the long-term reward consideration and informative guidance from state definition in RL models, which are particularly beneficial to efficient transit control. In particular, the proposed CAAC performs the best in stabilizing the system: it shows the smallest AWT and AOD values, while imposing more control interventions than MADDPG. This is mainly due to the fact that MADDPG does not distinguish its own contribution from that of others. Consequently, the reward in MADDPG is mainly determined by the agent's own control actions, leading to biased strategies. IAC can be considered a special case of CAAC, with the event critic component removed. Our results show that IAC is also less effective than CAAC, since the impact from other agents is completely ignored in IAC and therefore it suffers from the uncertainty of the system dynamics.

\begin{figure}[!htbp]
\centering
     \includegraphics[scale=0.30]{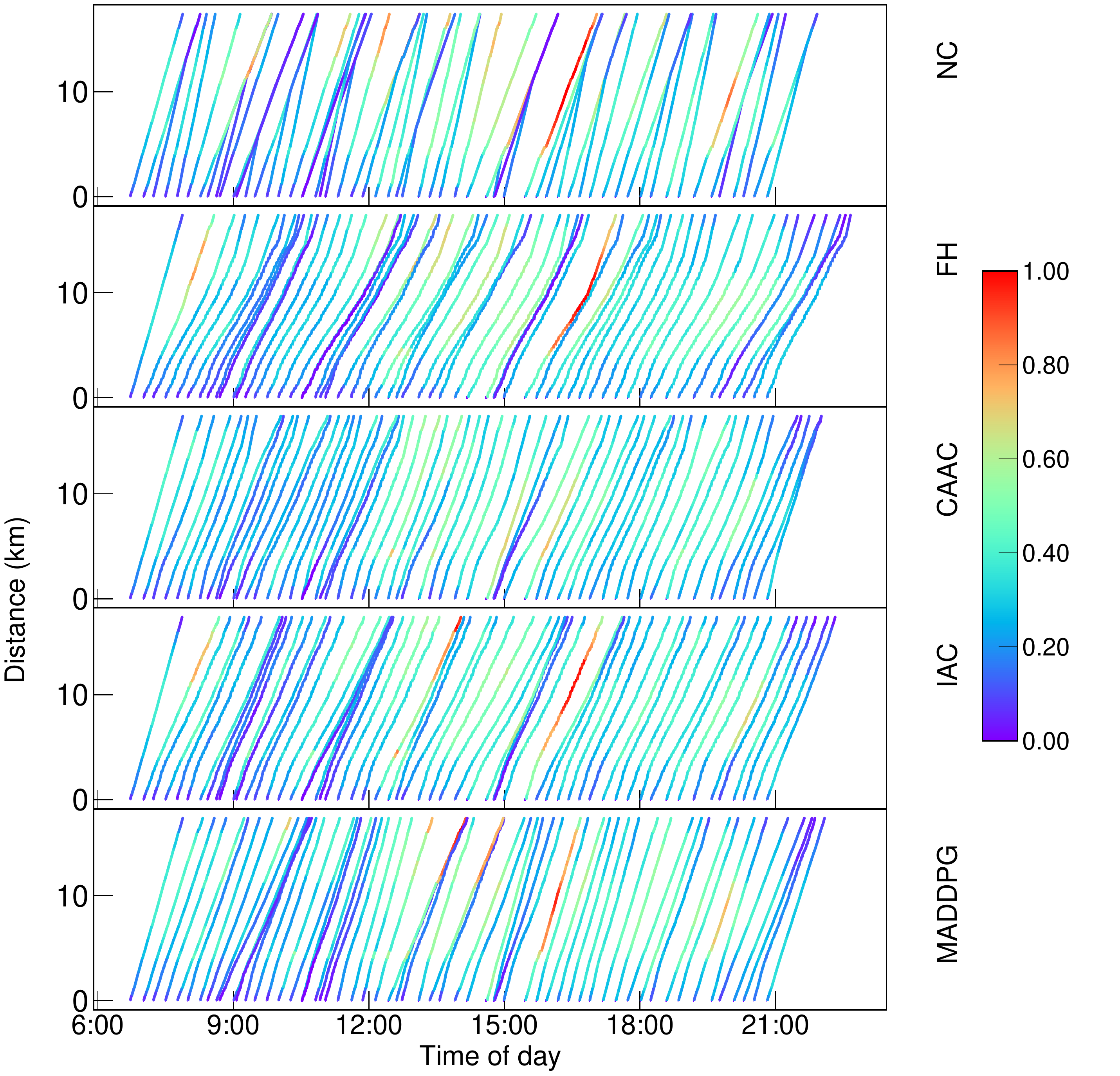}
  \caption{Bus trajectories under different holding strategies.}
   \label{fig:tr}
\end{figure}

Fig.~\ref{fig:tr} shows the trajectory plot colored by occupancy (i.e., number of onboard/capacity) under different control policies based on one random seed. As can be seen, the real-world operation provides a challenging scenario where it is hard to maintain headway regularity over the whole day without any intervention (i.e., NC). Notably, all MARL strategies seem to bring improvements at the upstream segment (i.e., 0-5 km). However, we see that CAAC clearly outperforms IAC and MADDPG in maintaining headway consistency over the downstream segment.

\begin{table*}[!h]
\begin{center}
\footnotesize
\begin{tabular}{ c c c c c}
\toprule
{Method} &\multicolumn{4}{c}{Performance on R2/R3/R4}  \\
\midrule
~ & AHT (sec) &$ \Delta$ AWT (sec) & $ \Delta$ AJT (sec) &$ \Delta$ AOD \\
\midrule

NC &   -/ - / -
&   625 / 757 / 723
&   1415 / 1389 / 1109
&   18.4 / 13.1 / 9.9\\

FH &   67/ 67 / 68
&   -87/ -70 / -36
&   +614 / +780 / +480
&   -5.1 / -2.4 / -2.5\\

CAAC &  42/ 40$^\ast$ / 41$^\ast$
&  \textbf{-166}/  \textbf{-161} /  \textbf{-127}
&   +251$^\ast$ / +336$^\ast$ / +192$^\ast$
&    \textbf{-9.9} /  \textbf{-5.9} /  \textbf{-4.9}\\

IAC &   40$^\ast$/ 42 / 42
&   -128/ -128 / -94
&   +254 / +360 / +202
&   -7.5 / -4.7 / -4.0\\

MADDPG &    \textbf{37}/ \textbf{35} / \textbf{36}
&    -132$^\ast$/ -145$^\ast$ / -111$^\ast$
&    \textbf{+202} /  \textbf{+223} /  \textbf{+130}
&   -7.7$^\ast$ / -5.0$^\ast$ / -3.7$^\ast$\\

\bottomrule
\end{tabular}
\caption{Execution performance when applying models trained on R1 directly on R2/R3/R4 without retraining. Model performance is evaluated using: 1) average holding time (AHT), and changes in---2) average waiting time ($\Delta$AWT), 3) average travel time ($\Delta$ATT), and 4) average occupancy dispersion ($\Delta$AOD)---w.r.t. the NC baseline. The best and second best results are highlighted in bold and with asterisk ($*$), respectively.}\label{table:test}
\end{center}
\end{table*}

\subsubsection{Execution on Untrained Routes (R2/R3/R4)}

When applying RL in real-world applications, one critical challenge is to ensure model transferability. As we adopt inductive graph learning in designing the event critic, we expect that trained model on one service can be also applied/generalized to other unseen services. To examine the transferability, we apply the models trained on R1 directly on R2/R3/R4 without retraining. Note that the configurations of R2/R3/R4 in terms of bus stop location, number of stops, and number of buses (i.e., agents) are different from that of R1. In addition, the default timetable and the demand pattern on R2/R3/R4 also differ substantially from those of R1. Thus, this experiment presents an unseen/unfamiliar environment for those RL models trained on R1. To better compare the models, we randomly scale demand with multiple random seeds and runs the model 10 times for each random seed.

Table~\ref{table:test} summarizes the overall performance of different models. For IAC, agents are trained in the environment ignoring the impact from other agents on the system. When transferred to another scenario, the performance greatly deteriorates due to the variation in dynamics from other agents. For MADDPG, it only partially considers actions from other agents given the uncertainty in agent states (i.e., activated or not). As a result, MADDPG performs worse on R2/R3/R4 than on R1. In contrast, the proposed CAAC shows superior transferability when applied to R2/R3/R4 in terms of system stability, maintaining comparable performance to that of R1. The above transferability analysis reinforces the point that, in an asynchronous setting, the training process clearly benefits from taking those undetermined events into consideration. As an example, CAAC demonstrates remarkable inductive power and transferability to unseen scenarios.


\section{Conclusion}

In this paper, we develop an asynchronous multi-agent reinforcement learning model to improve dynamic bus control and reduce bus bunching. In particular, we propose CAAC---a credit assignment framework to address the asynchronous issue in multi-agent control problems, which is often overlooked in previous studies. To effectively consider the impact/contribution of other agents, we design an inductive critic by combining an ego critic and an event critic. This design can be easily embedded in the actor-critic framework, and it turns out to be efficient and effective in policy training. Overall, our results demonstrate that the proposed CAAC framework is simple but very effective in characterizing the marginal contribution from other agents in bus holding as an ASMR problem. The graph neural network-based inductive critic offers additional generalization and inductive power, which allows us to apply a well-trained model directly on a new bus route.

There are several directions for future research. While the current model trains the agents with randomly scaled demand, we suspect that the performance of agents can be further improved using more informative/reasonable demand patterns. Another future research direction is to incorporate more traffic entities (e.g., traffic signals) with the inductive architecture and develop a more comprehensive traffic control policy such as signal priority for buses. Finally, the method can be generalized to other applications with similar spatiotemporal asynchronous patterns.

\section*{Acknowledgments}
This research is supported by the Fonds de Recherche du Qu\'{e}bec - Soci\'{e}t\'{e} et Culture (FRQSC) under the NSFC-FRQSC Research Program on Smart Cities and Big Data, and the Canada Foundation for Innovation (CFI) John R. Evans Leaders Fund.

\bibliographystyle{named}
\bibliography{ijcai21}

\end{document}